\documentclass[10pt,twocolumn,letterpaper]{article}

\usepackage{3dv}
\makeatletter
\@namedef{ver@everyshi.sty}{}
\makeatother
\usepackage{times}
\usepackage{epsfig}
\usepackage{graphicx}
\usepackage{amsmath}
\usepackage{amssymb}

\usepackage{pstricks}
\usepackage{caption}
\usepackage{subcaption}
\usepackage{pgfplots}
\usepackage{xspace}
\usepackage{layouts}
\usepackage{stmaryrd}
\usepackage{ulem}
\usepackage{fontawesome5}
\usepackage{float}
\usepackage[pagebackref=true,breaklinks=true,colorlinks,bookmarks=false]{hyperref}
\usepackage{pdfpages}

\threedvfinalcopy 

\def\threedvPaperID{118} 

\newcommand{\morphology}{\beta}
\newcommand{\poseSeq}{\Theta}
\newcommand{\pose}{\theta}
\newcommand{\globalSeq}{\Gamma}
\newcommand{\globale}{\gamma}
\newcommand{\timeSeq}{\mathcal{T}}
\newcommand{\timestamp}{\tau}
\newcommand{\motionRep}{\chi}
\newcommand{\latentVec}{z}
\newcommand{\loss}{\mathcal{L}}

\newcommand{\meshSeq}{M}
\newcommand{\mesh}{m}
\newcommand{\smseq}{\mathcal{F}}
\newcommand{\sm}{SMPL}

\renewcommand{\eg}{\textit{e.g.}}
\renewcommand{\ie}{\textit{i.e.}}

\ifthreedvfinal\fi
\begin{document}

\title{A Structured Latent Space for Human Body Motion Generation}

\author{Mathieu Marsot$^1$~~~
Stefanie Wuhrer$^1$~~~
Jean-S\'{e}bastien Franco$^1$~~~
Stephane Durocher$^2$\\
$^1$ \small{Univ. Grenoble Alpes, Inria, CNRS, Grenoble INP\thanks{Institute of Engineering Univ. Grenoble Alpes}, LJK, 38000 Grenoble, France}\\
$^2$ \small{University of Manitoba 66 Chancellors Cir, Winnipeg MB R3T 2N2, Canada}\\
{\tt\small firstname.lastname@inria.fr, stephane.durocher@umanitoba.ca}
}

\maketitle

\begin{abstract}
   We propose a framework to learn a structured latent space to represent 4D human body motion, where each latent vector encodes a full motion of the whole 3D human shape. On one hand several data-driven skeletal animation models exist proposing motion spaces of temporally dense motion signals, but based on geometrically sparse kinematic representations. On the other hand many methods exist to build shape spaces of dense 3D geometry, but for static frames. We bring together both concepts, proposing a motion space that is dense both temporally and geometrically. Once trained, our model generates a multi-frame sequence of dense 3D meshes based on a single point in a low-dimensional latent space. This latent space is built to be structured, such that similar motions form clusters. It also embeds variations of duration in the latent vector, allowing semantically close sequences that differ only by temporal unfolding to share similar latent vectors. We demonstrate experimentally the structural properties of our latent space, and show it can be used to generate plausible interpolations between different actions. We also apply our model to 4D human motion completion, showing its promising abilities to learn spatio-temporal features of human motion. Code is available at \href{https://github.com/mmarsot/A_structured_latent_space}{https://github.com/mmarsot/A\_structured\_latent\_space}.
\end{abstract}

\section{Introduction}

This work investigates learning a structured latent space to represent and generate temporally and spatially dense 4D human body motion, where a single point of a low-dimensional latent space represents a multi-frame sequence of dense 3D meshes. Recently, several works have proposed to learn such motion priors for 4D human body sequences of arbitrary motion by capturing information about pose changes over time~\cite{li2021taskgeneric,xu2021exploring,Lohit_2021_WACV}, in the case of fixed sequence duration. Here, we investigate an orthogonal scenario which models sequences of \textit{varying duration}, by considering motions sufficiently similar to allow temporal alignment. 

\begin{figure*}[t]
    \centering
   
    \includegraphics[width=\textwidth]{ 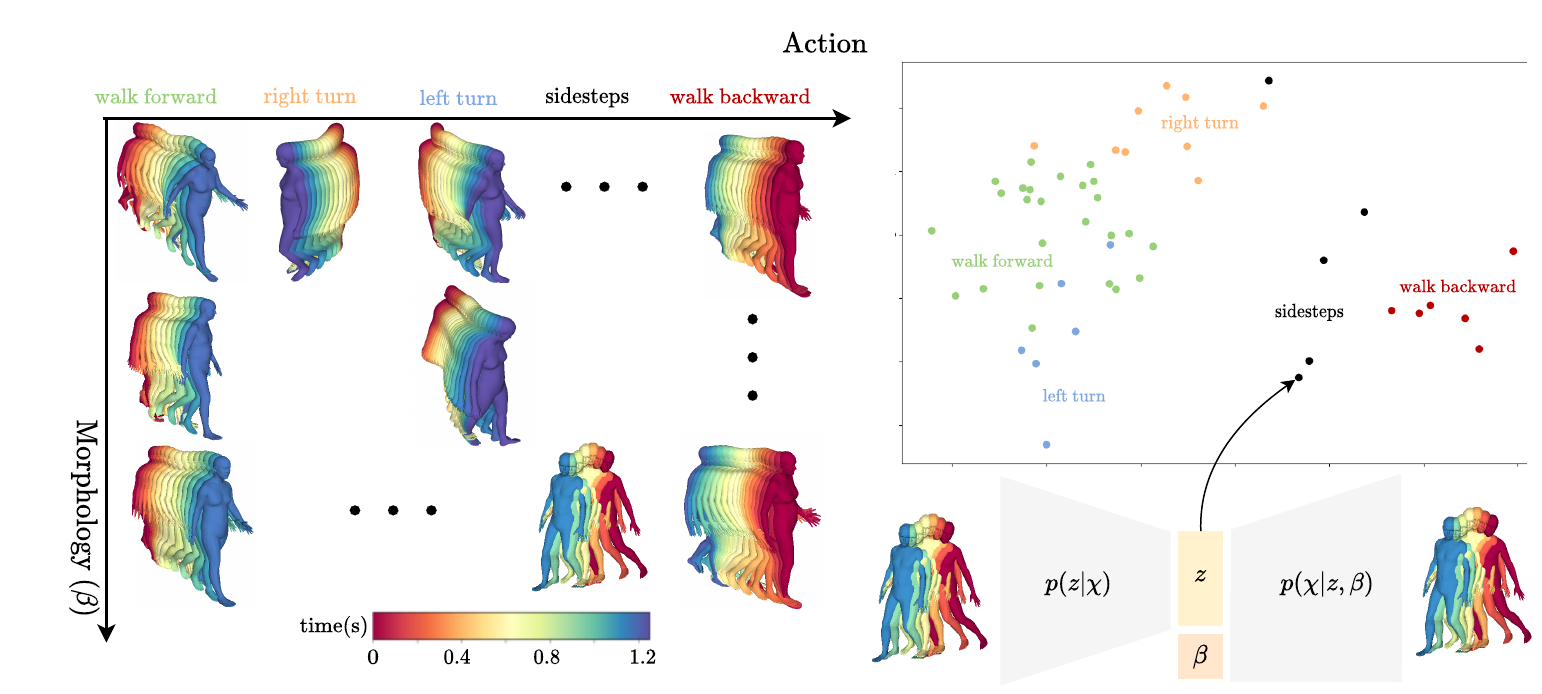}
    \vspace*{-4mm}
    \caption{We learn a latent \textit{motion space} from multi-frame 4D sequences. Left: Training sequences consist of different motions performed by different subjects (color-coded as shown in legend). Bottom right: Encoder-decoder architecture learns a latent motion space that encodes motion sequence $\motionRep$ into latent vector $\latentVec$; the decoder conditions $\latentVec$ on morphology $\morphology$. Top right: Structured latent space. Plot shows subset of 51 motions, manually labelled by action, in 2D projection of latent space. Actions form clusters.}
    \label{fig:data_pres}
    
\end{figure*}

Learning a generative model of 3D human motion of varying duration with structured latent space is of interest for a wide set of applications in computer vision and graphics, where a lightweight 4D representation translates to gains in information processing. By capturing a spatio-temporal motion prior, the model opens new directions for many completion tasks given temporally, geometrically sparse or incomplete inputs, as it allows to reason within a restricted plausible spatio-temporal solution space.

Learning this space is a difficult task with two major challenges. First, the model needs to capture the intertwined variations of different factors, \eg~morphology, global motion, body pose, and temporal evolution of the motion, and do so for motions that differ in duration. In particular, while it is known that morphology impacts the way a motion is performed~\cite{troje2002decomposing,physicsbasedshape2019}, it remains challenging to take this correlation into account during motion generation. Second, the amount of data that needs to be processed for training is large, as typical acquisition systems for dense human body motions produce $30-50$ frames per second, with each frame containing thousands of geometric primitives. 

To address these challenges, we take inspiration from two existing lines of work. The first studies temporally dense skeletal data, with the goal of generating skeletal human motion sequences that capture the temporal evolution of the global motion~\cite{troje08,sigal10,martinez17}. These do not address dense surfaces. The second line of work represents realistic 3D human body surfaces in a low-dimensional shape space~\cite{Anguelov05,Pons-Moll2015}, but do not consider the temporal dimension. 

We combine the advantages of both in a data-driven framework that learns a latent motion representation, which allows to simultaneously represent temporal motion information and detailed 3D geometry at every time instant of the motion. The learning uses multi-frame sequences as input and output. Inspired by works on morphable body models~\cite{troje2002decomposing,Allen2003}, we align the training sequences both temporally and spatially, which leads to comparisons at corresponding instances of the motion and anatomically corresponding points. In particular, we consider motions whose duration vary significantly while geometrically similar enough to allow for temporal alignment, performed by actors in minimal clothing to allow for effective spatial alignment. 

In our experiments, we consider motions during which the hip performs a cycle, as this includes common motions such as walking and running, and generalizes to more complex motions such as dancing or jumping jacks, while imposing no constraints on the arm movements. 
The resulting latent space is verifiably structured, and allows to generate plausible interpolations between different types of locomotion that outperform linear and per-frame interpolation baselines.
As illustrated in Fig.~\ref{fig:data_pres}, our motion space also learns the interaction between morphology and motion, as generating motions with the same point in latent space conditioned on different representations of morphology leads to motion differences that confirm findings in prior studies conducted on sparse motion data~\cite{troje2002decomposing}.

Our model can serve as prior to complete both spatially and temporally sparse sequences. Given as input unmatched and temporally incoherent point clouds sparsely sampled in space or time, accurate complete 4D reconstructions are obtained. For spatio-temporal completion, our method outperforms state of the art motion priors that encode human motion sequences of fixed duration~\cite{li2021taskgeneric,xu2021exploring} with sufficient temporal samples, 
in spite of being trained on significantly less data. It also outperforms a state of the art spatial completion baseline when few samples are available~\cite{zhou2020reconstructing}.

In summary, we make the following major contributions. First, we present a latent motion space that allows representing and generating multi-frame sequences of dense 3D meshes of varying duration, which accounts for interaction between morphology and motion. Second, we demonstrate that this latent space is structured: similar motions form clusters, and linear interpolation in latent space outperforms baselines. Third, when using our motion space as prior, we outperform state of the art for the application of motion completion from sparsely sampled data in space or time. 

\section{Related Work}
\label{sec_related}

The vast literature on generation of human models and motions can be roughly divided into three categories. \textit{Temporally dense} encompasses methods that learn the structure of human motion on a representation that is sparse in 3D space. \textit{Spatially dense} encompasses methods that generate realistic 3D human models without treating long-term motion or dynamic effects. \textit{Full 4D} methods combine long-term motion models with dense 3D shapes per frame.

The first two lines of work have been studied for the past two decades. Studies on temporally dense human motion models proposed different data-driven methods to synthesize motion patterns of skeletal representations or sparse marker positions~\eg~\cite{rose98, troje02, troje08, Holden2016, Lohit_2021_WACV}. These works effectively learn the structure of human motion over durations of multiple seconds.
Studies on spatially dense human models proposed a variety of data-driven methods to synthesize geometrically detailed 3D models~\eg~\cite{Allen2003, Anguelov05, Neophytou2013, Pishchulin2017, loper2015smpl}. Some models have been extended to learn soft-tissue deformations~\cite{Pons-Moll2015, Loper14, santesteban2020softsmpl}. Recent works in this area leverage deep learning techniques, and can decouple variations due to different factors~\eg~\cite{jiang20, cosmo20, zhou20unsupervised} or include hands, faces and soft-tissue deformation,~\eg~\cite{xu20}. These works generate realistic and geometrically detailed 3D human models.

Over the past few years, a number of works proposed studying 4D human motion data that is densely sampled in space and time. Some work aims to generate dense 3D human motion from sparse MoCap~\cite{Anguelov05, Loper14, mahmood2019amass, habermann2021} or 2D video data~\cite{kanazawa19, zhang19}. Given as input marker points or a 2D image per frame of the motion, these works reconstruct dense 4D motion data. Of particular interest for our work is that statistical body models learned on static data have been fitted to MoCap data, providing a large corpus of semi-synthetic dense 4D data~\cite{mahmood2019amass}. This provides the community access to a large 4D dataset, which we leverage in our work.

The works most related to ours learn spatially and temporally dense 4D motion models of bodies in a data-driven way.
The first work to tackle this problem~\cite{kuznetsova13} combines two linear models: one capturing dense static 3D shape data and one capturing the motion of MoCap markers. The two linear models are coupled based on semantic parameters including weight and height, which allows generating 4D human motion sequences. Inspired by this idea, our model learns a non-linear model from 4D data, which includes both morphology and motion. We show experimentally that our model generalizes better than a linear one.

With 4D data becoming increasingly available in recent years, a number of studies propose data-driven methods trained on 4D data. First methods including~\cite{Akhter2012, boukhayma18, regateiro19,regateiro2021deep4d} train on either a single motion sequence or multiple sequences showing the same subject performing different motions. A recent work that studies motions of a single subject proposes a deep latent variable model for 4D human motion synthesis~\cite{Ghorbani2020} to model the probabilistic character of motion. 

Recently, 4D motion priors of different subjects performing different motions have received considerable attention. One line of work uses implicitly defined surfaces over time to learn from raw 4D sequences~\cite{occupancyFlow,jiang2021learning}, and successfully process human motion data. However, the high dimensionality of the 4D data constrains the sequences to few frames. 

To consider longer temporal spans, other works build motion priors from sequences of pose parameters of template aligned meshes. These works include methods that consider a set of labeled actions to learn motion generation based on action labels~\cite{ACTOR:ICCV:2021} and methods that model motion as a sequence of transitions between poses~\cite{rempe2021humor,lee2010motion}. Most similar to our work are methods that build motion priors of unlabeled 4D human motion data~\cite{li2021taskgeneric,xu2021exploring}. These methods consider motions of a fixed duration and encode them in a motion space, which captures information about pose changes over time. In contrast, we investigate learning a motion space for 4D sequences of varying duration. We demonstrate experimentally that our motion space outperforms~\cite{li2021taskgeneric} and \cite{xu2021exploring} for motion completion.

\section{Generative model of multi-frame sequences}
\label{generative}

Two previously identified major challenges need to be tackled in our model: first the very large dimensionality of the problem as is concerns temporally dense sequences of dense 3D meshes; second the modeling of intertwined variations in the generation of 4D sequences, between subject shape, morphology, motion, and temporal unfolding.

To address them, we first need to ensure that we produce a compact and structured motion representation. Our general strategy for this is to extend the static shape space representations (\eg~SCAPE) to the spatio-temporal domain, with a similar low-dimensionality characteristic, as detailed in Section~\ref{generative-rep}.
Second, we articulate our data-driven strategy around an encoder-decoder architecture (Section~\ref{sec:archi}). Notably, to explicitly model the interaction between morphology and motion, we choose to condition the motion generation on a representation of morphology. Third, we build our experimental demonstration in a use case that benefits from these choices, focusing our effort on a database of 4D human motion sequences that perform a cyclic motion of the hip joint. This allows to evidence the intended behaviour for this space, which is to group similar locomotions (\eg~all walking motions) in clusters. 
Section~\ref{generative-train} explains how the model is trained.

\subsection{Representation of motion sequences}
\label{generative-rep}

Fig.~\ref{fig:archi} (top left) shows our representation for 4D sequences. A 4D human motion sequence is parameterized by a single point $\latentVec$ in motion space and a identity parameter $\morphology$ representing the morphology of the moving person.

\textbf{Anchor frames} To represent motion data, we align an unstructured spatio-temporal motion signal. 
Temporally, we uniformly sample $n$ frames from the motion signal, which we call anchor frames in the following. These anchor frames allow representing motions of various duration with the same number of frames.  
Spatially, we build on 3D morphable body models to align the frames \eg~\cite{Neophytou2013,Pishchulin2017,loper2015smpl}. These models represent static 3D human body surfaces using a common mesh template. This results in $n$ aligned anchor meshes, making motion comparison practical.

\textbf{Representing temporal evolution} The resulting anchor mesh sequence $\meshSeq=[\mesh_1,\ldots,\mesh_n]$ does not represent the temporal evolution of a motion. The temporal sampling causes an information loss, as it is invariant to similar motions with different temporal unfolding like walking and running. Therefore, we associate to anchor mesh $\mesh_i$ a timestamp $\timestamp_i$, and call the timestamp vector $\timeSeq = [\timestamp_1,\ldots,\timestamp_n$]. The representation $[\meshSeq, \timeSeq]$ is high-dimensional. To simplify processing and disentangle the influence of morphology on motion, we leverage 3D morphable body models that decouple the influence of morphology and pose. By holding morphology constant over $\meshSeq$, we can represent each $\mesh_i$ using parameter vectors for morphology $\morphology$, pose $\pose_i$, and global translation $\globale_i$. While any decoupled static model can be used,~\eg~\cite{jiang20, cosmo20, zhou20unsupervised}, in our implementation we chose the commonly used \sm~model~\cite{loper2015smpl} as the AMASS dataset~\cite{mahmood2019amass} is parameterized by \sm. We denote the model function by $\sm$ such that $\mesh_i = \sm(\theta_i,\globale_i,\morphology)$ and thus $\meshSeq = [(\sm(\theta_0,\globale_0,\morphology),\ldots,\sm(\theta_n,\globale_n,\morphology)].$
By denoting the pose and global translation vectors by $\poseSeq=[\pose_1,\ldots,\pose_n]$ and $\globalSeq = [\globale_1,\ldots, \globale_n]$, respectively, $[\poseSeq,\globalSeq,\morphology,\timeSeq]$ is a low dimensional representation of $[\meshSeq,\timeSeq]$. To retain variation in global displacement (\eg walking backward or forward) and temporal evolution (\eg walking or running), we model $\globalSeq$ and $\timeSeq$ in the multi-frame sequence representation. $\timestamp_i$ allow to place freely and on any time span length the anchor meshes, thereby allowing to represent motions with various duration using a constant number of meshes.

\textbf{Notation} To emphasize the difference between motion and morphology parameters, we denote  $\motionRep = \left[ \poseSeq, \globalSeq, \timeSeq \right]$ the motion parameters and introduce function $\smseq$ such that $[\meshSeq,\timeSeq] = \smseq([\motionRep,\morphology])$. As pre-processing for training, we map a raw motion sequence to the \sm~mesh template using existing solutions~\cite{Yang2016,mahmood2019amass}. Let $\sm^{-1}$ denote the mapping function which associates a single raw motion frame to its representation parameters $\pose,\globale,\morphology$.

\textbf{Numerical representation} In practice, we represent $\morphology$ and $\globalSeq$ as in \sm. Pose features $\poseSeq$ are joint rotations of a skeleton, represented by a continuous 6D rotation~\cite{zhou_continuity_2019} that was shown to outperform other rotation representations when training neural networks. 

\subsection{Architecture}
\label{sec:archi}
\begin{figure}
    \centering
    \scalebox{1}{\includegraphics[width=0.45\textwidth]{ 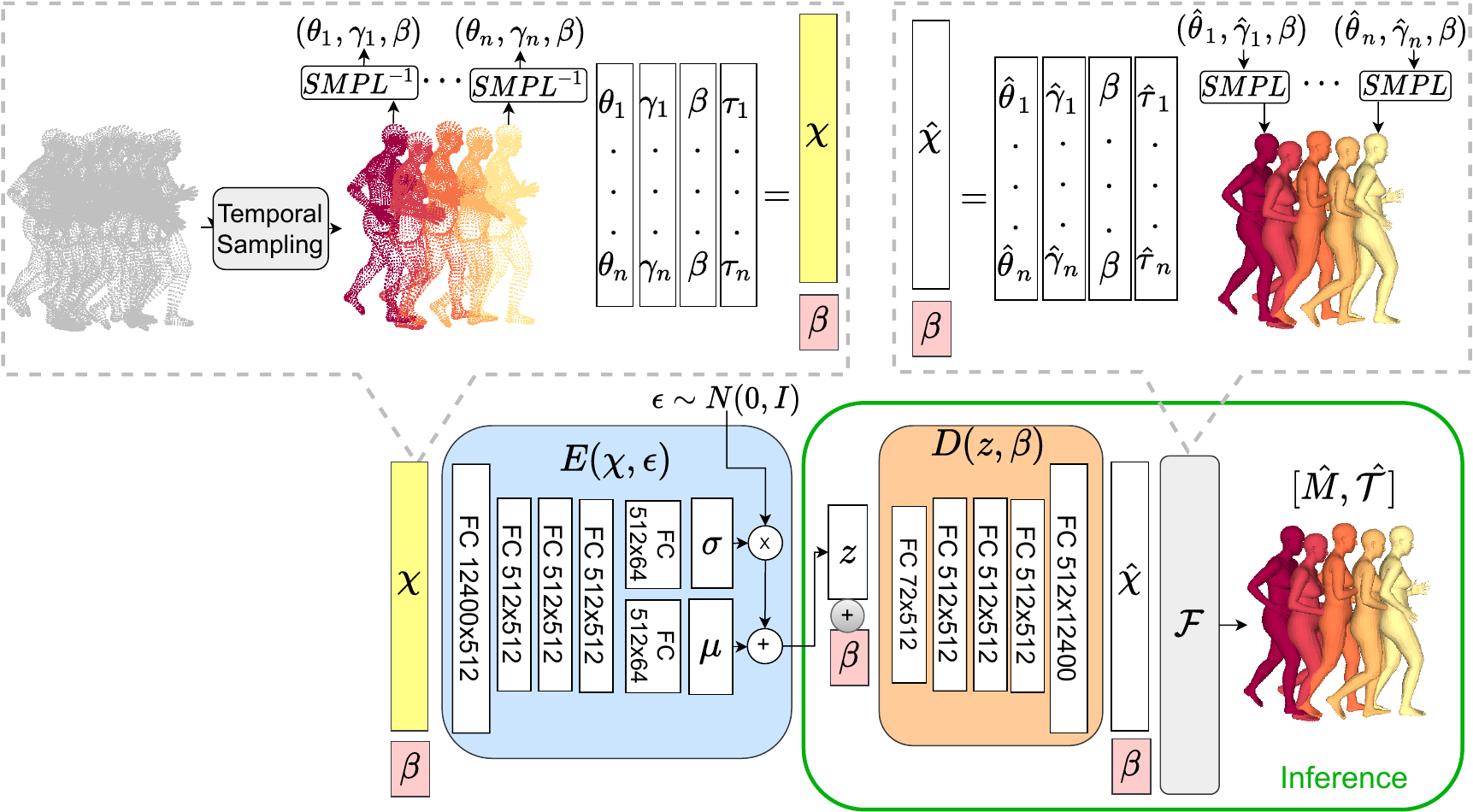}}
   
    \caption{Overview of motion representation and architecture. Top: representation. Left: pre-processing during training samples $n$ anchor frames and extracts per-frame representations of pose $\pose$, translation $\globale$ and morphology $\morphology$ with their timestamp $\timestamp$ to obtain motion representation $\motionRep$ and morphology $\morphology$. Right: illustration of the function $\smseq$. Bottom: our architecture consists of a probabilistic encoder $E$ and a decoder $D$, and learns a mapping from $\motionRep$ to a single latent vector $\latentVec$. At inference time, $D$ conditions $\latentVec$ on $\morphology$ to generate sequence features $\hat \motionRep$ (green box). \looseness=-1}
 
    \label{fig:archi}
\end{figure}

To learn the interaction between morphology and motion patterns, we condition motion generation on $\morphology$ using an architecture based on conditional variational auto-encoders (CVAE)~\cite{sohn2015learning}, as shown in the bottom of Fig.~\ref{fig:archi}. Our architecture encodes motion vector $\motionRep$ into a low-dimensional latent vector $\latentVec$, and $\morphology$ is used as condition for the decoder, thereby allowing to capture dependencies between $\motionRep$ and $\morphology$.
We assume $\latentVec$ and $\morphology$ to be independent and learn a disentangled representation. Therefore, the encoder models posterior distribution $p(\latentVec|\motionRep)$, and is not conditioned on $\morphology$. 

The encoder outputs are interpreted as mean $\mu$ and standard deviation $\sigma$ of the posterior distribution of the latent space. The corresponding latent vector $\latentVec$ is sampled as $\latentVec = \mu + \epsilon \times \sigma$, with $\epsilon \sim \mathcal{N}(0,1)$. We denote the probabilistic encoding function by $E : \motionRep,\epsilon \mapsto \latentVec$, and the decoding function as $D: \latentVec,\morphology \mapsto \hat\motionRep$.
The decoder takes $(\latentVec,\morphology)$ as input, and outputs $ \hat{\motionRep} = [\hat{\poseSeq}, \hat{\globalSeq}, \hat{\timeSeq}]$ which are converted back to a sequence of timestamped anchor meshes $[\hat\meshSeq, \hat{\timeSeq}]=\smseq(\hat\motionRep,\morphology)$. To go from a reconstructed sequence $\hat\meshSeq $ to a temporally continuous motion, we assume constant motion between anchor meshes.  

\subsection{Training}
\label{generative-train}

The network is trained with a reconstruction term to minimize the difference between the input and output vectors, and a regularization term to constrain the latent variables to follow a known prior distribution. The training is divided into two phases. First, we consider a reconstruction loss on $\motionRep$ to allow for fast and memory efficient initialization. Second, we replace it by a loss computed directly on the sequence of anchor meshes $\meshSeq$ in $\mathbb{R}^3$. 

\textbf{Reconstruction loss on $\motionRep$}
The standard reconstruction term would be $(\hat{\motionRep}-\motionRep)^2$. To balance the influence of the different types of information captured by $\motionRep$, we divide this loss into three terms operating on pose $\loss_{pose} = ( \poseSeq - \hat{\poseSeq} )^2$ translation $\loss_{trans} = (  \globalSeq - \hat{\globalSeq} )^2$, and time  $\loss_{time} = ( \timeSeq - \hat{\timeSeq} )^2$. This gives a total reconstruction loss
\begin{equation}
    \label{eq:lsmpl}
    \loss_{rec} = \omega_{pose}\loss_{pose}+ \omega_{trans}\loss_{trans} + \omega_{time}\loss_{time},
\end{equation}
where $\omega_{pose}, \omega_{trans}$ and $\omega_{time}$ are the respective weights of the partial reconstruction losses.
To minimize $\loss_{rec}$, we use adaptive weights to trade off the relative influence of $\loss_{pose},\loss_{trans}$ and $\loss_{time}$~\cite{pmlr-v80-chen18a}, which do not have the same order of magnitude. Adaptive weights are initialized at 1.0 and updated automatically during training, which ensures that the partial losses are decreasing in similar proportions. 

\textbf{Reconstruction loss in 4D}
The second reconstruction loss is $\loss_{spatial} = (\meshSeq-\hat{\meshSeq})^2$, where $\meshSeq$ denotes the 3D coordinate vector or the anchor mesh sequence, resulting in the $4D$ reconstruction term 
\begin{equation}
    \label{eq:l4d}
    \loss_{rec4D} = \omega_{spatial}\loss_{spatial} + \omega_{time}\loss_{time},
\end{equation}
where $\omega_{spatial}$ is an adaptive weight. 

\textbf{Regularization loss}
The regularization term is the squared Kullback-Leibler (KL) divergence between the learned posterior distribution $\mathcal{N}(\mu,\sigma)$ of the latent variable $\latentVec$ and a normal prior distribution $\mathcal{N}(0,1)$, denoted $\loss_{KL}$.\looseness=-1

\textbf{Optimization}
A common problem when training VAEs is the weighting of the regularization loss versus the reconstruction loss. We use a fixed weight $\omega_{KL}=0.01$ to trade off these losses. The training optimizes first 
\begin{equation}
 \loss_{init} = \loss_{rec} + \omega_{KL} \loss_{KL}
\end{equation}
and subsequently
\begin{equation}
 \loss = \loss_{rec4D} + \omega_{KL} \loss_{KL}.
 \vspace*{-1mm}
\end{equation}


\section{Evaluation}
\label{sec_evaluation}

This section presents comparisons to baselines. We investigate the structure of the learned latent space by visualizing labeled motion sequences in latent space and by linearly interpolating between pairs of input motion sequence. Finally, we demonstrate that the proposed model learns information on the interaction of morphology and motion by visualizing the motion changes caused by changing $\morphology$ for a fixed point $\latentVec$. Implementation details, a study of the influence of the latent space dimension and regularisation, and video visualizations are in supplementary material.

\subsection{Data}
\label{sec_data}

We automatically extract motion sequences during which the hip performs a cycle from a dataset by comparing all subsequences to a set of 4D template motions using dynamic time warping~\cite{berndt1994using} as distance. Subsequences are considered if this distance is below a threshold. As post-processing, we prune segments with a duration above $3s$ or below $0.3s$. We manually generate two 4D template motions as gait cycles starting with the left and right foot. 

We experiment with AMASS~\cite{mahmood2019amass} and Kinovis~\cite{Yang2016} datasets. AMASS regroups a set of MoCap recordings and fits \sm~to all data. When splitting AMASS into training and test sets, we treat all sequences of the same MoCap dataset as one entity. For training, our cropping results in 12085 sequences corresponding to $\approx 4.5h$ of motion. We call the extracted test set \textit{AMASS test set}. Kinovis dataset contains 4D motion sequences from a multi-view platform and allows to evaluate the generalization of our model to densely captured 4D data. We consider all walking and running sequences, pre-process the data by fitting \sm~before extracting cyclic hip motions, and call this dataset \textit{Kinovis test set}. Details are in supplementary material. 

\subsection{Comparison to baseline models}

We compare our model to two baselines~\wrt the reconstruction error 
$\frac{1}{nk}(\meshSeq - \hat\meshSeq)^2,$
with $\left[\hat\meshSeq,\hat\timeSeq\right] = \smseq(\hat{\motionRep}, \morphology) =\smseq(D(E(\motionRep,\epsilon),\morphology),\morphology)$, where $n$ is the number of anchor frames and $k$ the number of vertices per frame. The first baseline applies a linear principal component analysis (PCA) to our representation $\left[ \motionRep , \morphology \right]$, thereby evaluating the value of using a non-linear model. PCA has access to morphology information when projecting the motion representation to latent space, and reconstructs both $\hat{\motionRep}$ and $\hat{\morphology}$. To provide a fair comparison, we consider the original $\morphology$ instead of $\hat{\morphology}$ in PCA reconstructions and set the PCA latent dimension to $dim(\latentVec)+dim(\morphology)$ with $dim(\latentVec)=64$ and $dim(\morphology)=8$. The second baseline considers our model after optimizing $\loss_{init}$ only, which operates on skeleton representations, thereby evaluating the value of learning from data that is densely sampled in space. 

\begin{figure}
    \begin{center}
        \scalebox{0.7}{\input{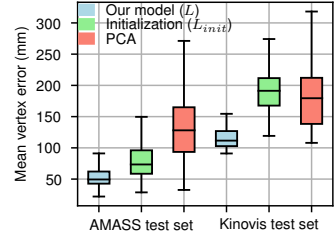}}
    \end{center}
    \vspace{-8mm}
    \caption{
	Comparison to baselines \wrt reconstruction error. Our model (blue) outperforms a linear PCA baseline (red) and a baseline that considers spatial sampling at skeleton level (green). Boxes follow \cite{tukey1977box}.}

    \label{fig:gen1}
\end{figure}

Fig.~\ref{fig:gen1} shows reconstruction errors for the different models. While PCA provides low reconstruction errors, these are further improved using our model. Our model also improves over its initialization, which shows that considering densely sampled data significantly impacts performance.

\subsection{Motion space structure and interpolation}

\begin{figure*}[ht]
    \centering
		\scalebox{0.9}{\begin{tabular}{cc}
		    (a) Interpolation of duration  & (b) Interpolation of global displacement \\
			\includegraphics[width=0.5\textwidth,height=30mm]{ 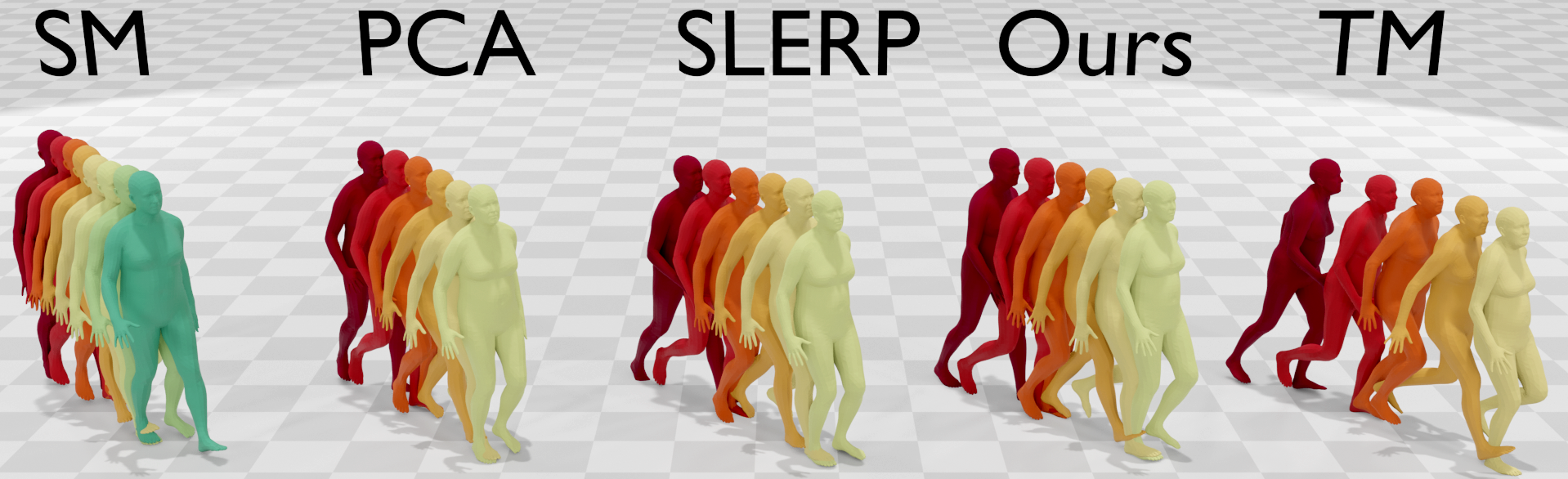} & \includegraphics[width=0.5\textwidth,height=30mm]{ 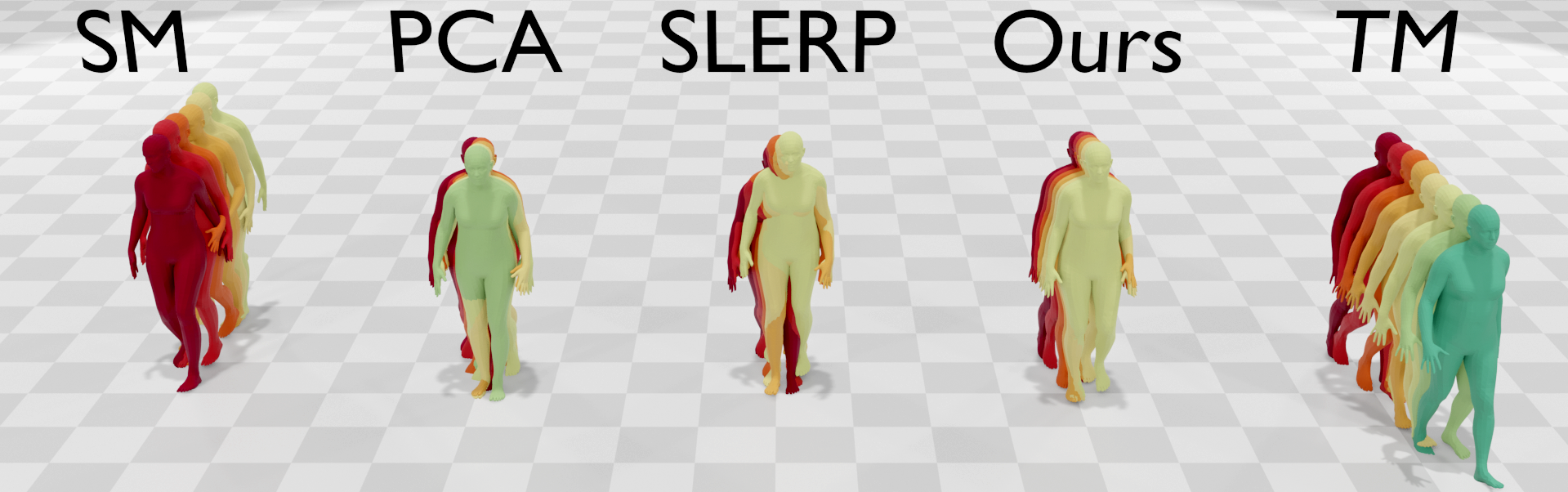} \\
			(c) Interpolation of pose (global)  & (d) Interpolation of pose (localized) \\
			\includegraphics[width=0.5\textwidth,height=30mm]{ 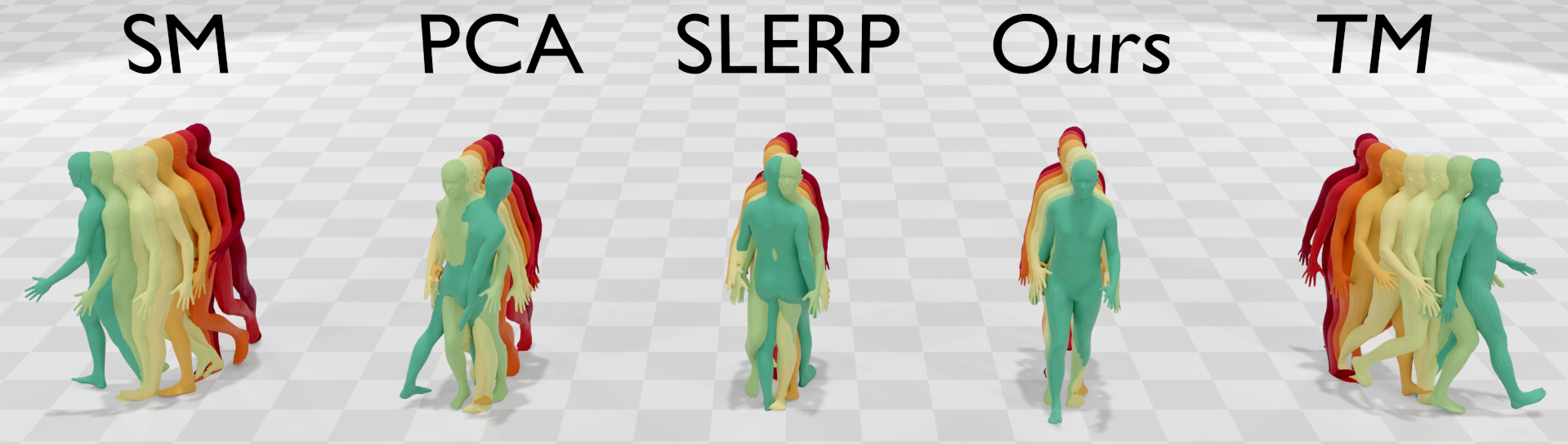} & \includegraphics[width=0.5\textwidth,height=30mm]{ 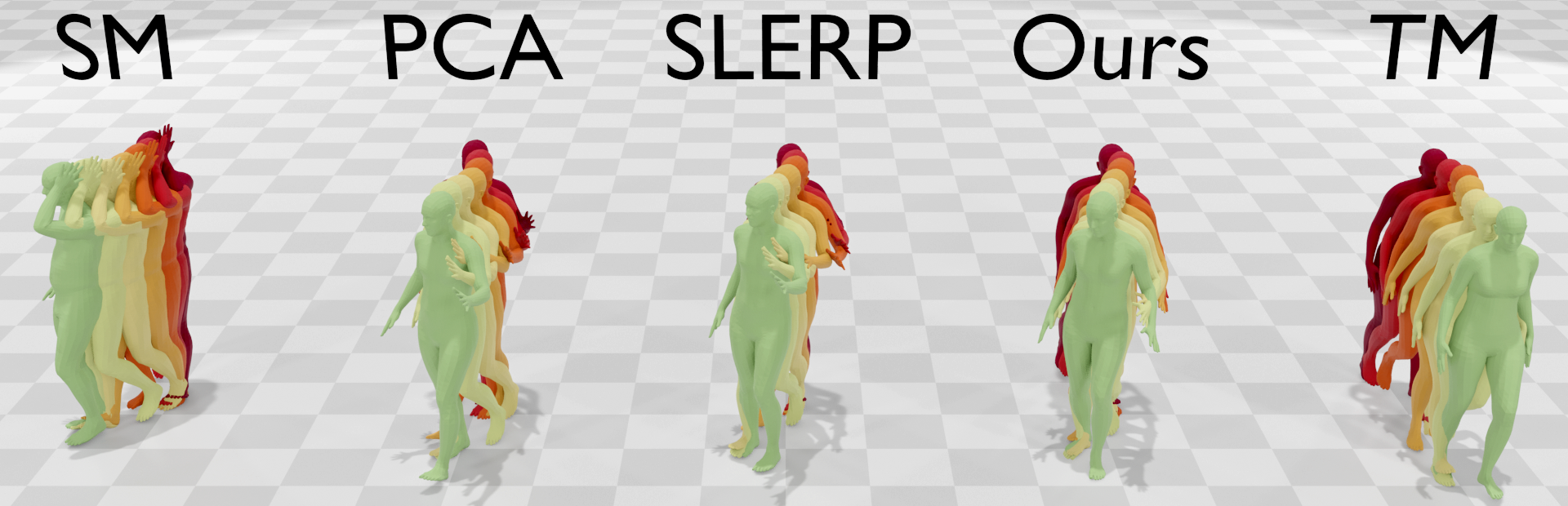} \\
			\multicolumn{2}{c|}{\includegraphics[width=0.3\textwidth,height=5mm]{ 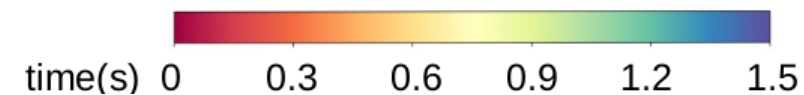}} \\
		\end{tabular}}
	
    \caption{Linear interpolations in latent motion space. 
		Each figure left to right : starting motion, PCA interpolation, SLERP interpolation, our interpolation, and target motion. Sequence models are rendered with a color-coded frame time. \textbf{(a)}  Running \& walking. \textbf{(b)}  Walking backward \& forward. \textbf{(c)}  Left \& right turn. \textbf{(d)}  Walk \& walk carrying an object on the head.
    All interpolations with our model are plausible, while baselines fail in (b) and (c).}
    \label{fig:interpolation}
    
\end{figure*}

Fig.~\ref{fig:data_pres} illustrates that our model learns a latent space in which sequences of similar actions are clustered. For the purpose of visualization, we labeled 51 motions by actions and assigned a unique color per action. These motions are then encoded into latent space, which is linearly reduced to two dimensions. Points of the same action form clusters.

This structured latent space can be exploited to generate plausible interpolations between input motions using linear interpolation. Given start and target motion sequences as input, we encode them as $(\latentVec_s, \morphology_s)$ and $(\latentVec_t, \morphology_t)$, and generate interpolating motion sequences by decoding $((1-k)\latentVec_s + k\latentVec_t, (1-k)\morphology_s + k \morphology_t)$ at intermediate position $k \in [0,1]$. 

We compare our results to two baselines. The first uses the PCA model from the previous section
and linearly interpolates in PCA space. This comparison, called PCA, evaluates the value of using a non-linear model. The second baseline operates per anchor frame and interpolates linearly between the global displacements, time stamps and morphology parameters, and with spherical linear interpolation~\cite{shoemake1985animating} (SLERP) between skeletal poses. This comparison, called SLERP, evaluates the value of learning a motion model instead of operating independently per-frame. For all interpolations, visualizations show $k=0.5$.
In the following, we interpolate between sequences that differ in each of the factors encoded in $\motionRep$. 

\textbf{Interpolating sequences of different duration}
To inspect temporal information learned by our model, we interpolate between a running and a walking motion. For our model, the duration of the intermediate sequences monotonically decreases when going from running to walking, and the intermediate sequences are realistic as shown in Fig.~\ref{fig:interpolation}(a), showing that our motion space has captured information on the temporal evolution $\timestamp$. PCA and SLERP baselines also lead to plausible interpolations.

\textbf{Interpolating sequences of different global displacement}
To inspect global displacement, we interpolate between a forward and a backward walk. Our intermediate sequence corresponds to a really small step, shown in Fig.~\ref{fig:interpolation}(b). There were no steps this small in the training set. 
PCA and SLERP baselines fail to interpolate global translation realistically, resulting in foot skating. 

\textbf{Interpolating sequences of different pose}
To inspect the learned information of pose, we consider global and articulated pose separately. First, we interpolate between sequences of turning left and turning right while walking, exhibiting mostly global pose change. The intermediate sequences using our model gradually change from a left to a right turn as shown in Fig.~\ref{fig:interpolation}(c). PCA and SLERP baselines fail due to the ambiguity when interpolating between opposite rotations, while our model leverages spatio-temporal information to alleviate this ambiguity. 
Second, we interpolate between walking and walking while carrying an object on the head, exhibiting mostly articulated pose change. The intermediate sequence with our model results in realistic intermediate positions for the arms, gradually elevating them to head level as shown in Fig.~\ref{fig:interpolation}(d). 
Both baselines lead to plausible interpolations.

In summary, while our model generates visually plausible interpolations for all parameters encoded in $\motionRep$, both baselines exhibit failure cases in some scenarios, which shows the value of learning a non-linear 4D motion model.

\subsection{Interaction between morphology and motion}

\begin{figure}
{\includegraphics[width=0.45\textwidth]{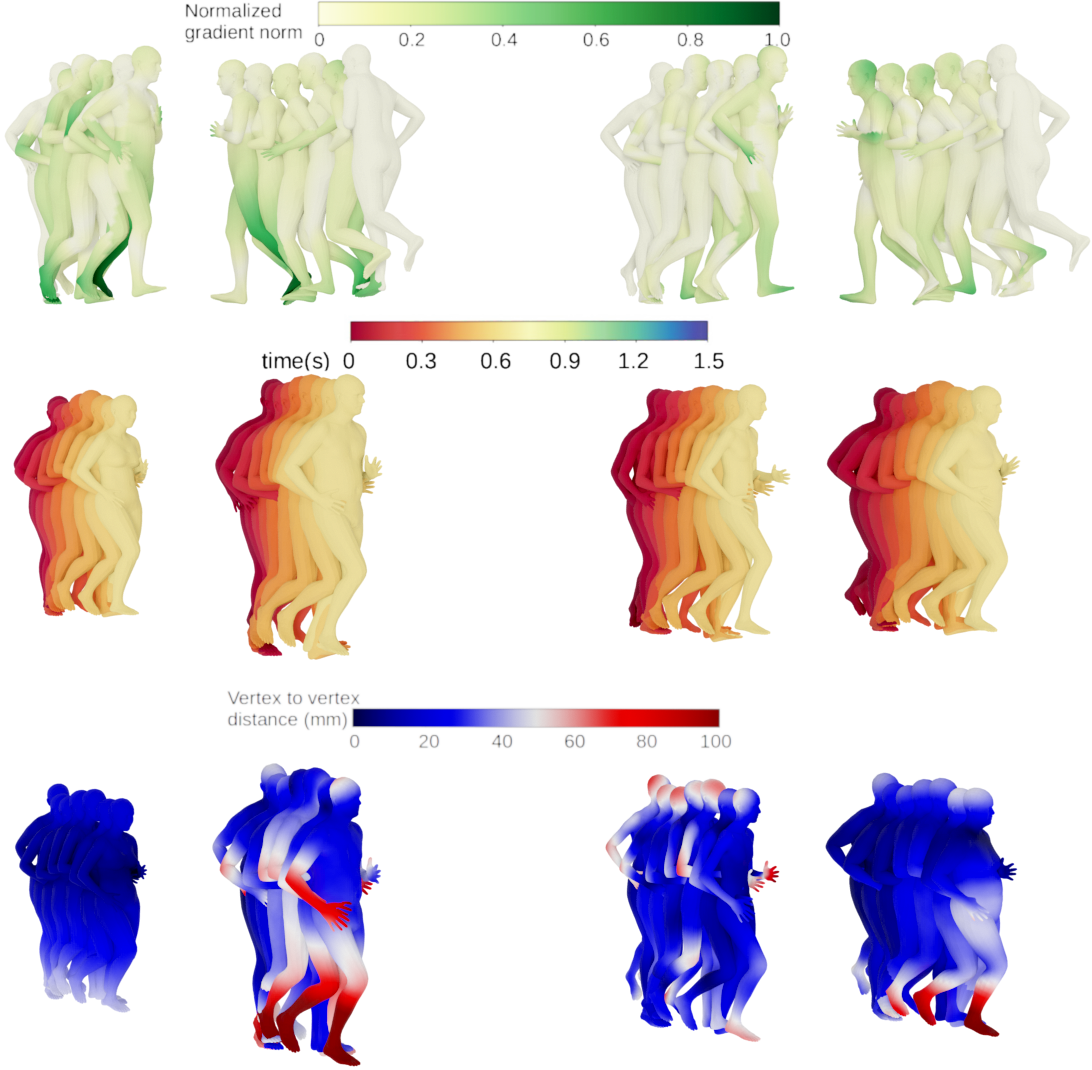}}
\vspace{-3mm}
\caption{Interaction between morphology and motion on $1^{st}$ (left) and $2^{nd}$ (right) principal components of $\morphology$. Top: visualization of our decoder's normalized gradient~\wrt $\morphology$. 
Middle: our inferences with fixed latent motion vector and $\morphology$ taken at $\pm 3$ std. deviations.
Bottom: baseline per-frame motion transfer using \sm~for same fixed motion and $\morphology$ taken at $\pm 3$ std. deviations, color coded by per-vertex distance to our result. 
Our learnt correlation has significant impact on motion, which differs up to $10cm$ from baseline.}

\label{fig:st1}
\end{figure}

To examine the influence of morphology $\morphology$ on 4D motion $\motionRep$, we consider a fixed jogging motion represented by $\latentVec^*$ in motion space and visualize $\motionRep$ when setting $\morphology$ to $\pm 3$ standard deviations along the first and second principal components. To understand the subtle motion differences, we further visualize the spatio-temporal gradient $\frac{\partial D(\latentVec^*, \morphology)}{\partial \morphology}$ at $\morphology = 0$,~\ie~we look at the gradient learned by the decoder \wrt~morphology at the mean shape.

We compare our result to a baseline that uses the initial pose parameters and $\beta$ to reconstruct a dense 3D body model using \sm~per frame. This evaluates the influence of learning the interaction between morphology and motion.

Fig.~\ref{fig:st1} shows the impact of the first (left) and second (right) principal components of $\beta$. The top row shows a color coding of the gradient learned by our decoder \wrt~$\morphology$ on the 4D sequence, and the middle row shows the corresponding 4D motions obtained by our model. The bottom row shows the result of the baseline color-coded by the distance to the result of our model. %
Changing the first principal component impacts perceived gender. For our model, this changes the 4D motion on the right shoulder and left hip, in agreement with prior studies showing that shoulder sway and hip motion are statistically gender related~\cite{troje2002decomposing}. %
Changing the second principal component leads to perceived weight change. For our model, this impacts the 4D motion at the right arm, head and neck. The spatio-temporal areas affected by our motion model are the ones where the baseline leads to significantly different results with up to $10cm$ distance.
This shows that our model learns meaningful interactions between morphology and motion.

\section{Application to motion completion from spatio-temporally sparse input}
\label{sec:completion}

This section applies our model to spatio-temporal completion, which has applications ranging from the registration of a raw spatio-temporally densely scanned 4D sequence over computing realistic in-betweenings for a set of frames sparsely sampled in time to completing full human body motion from a sparse set of MoCap markers.

\subsection{Completion methodology}

\begin{figure}
    \centering
    \includegraphics[width=0.5\textwidth]{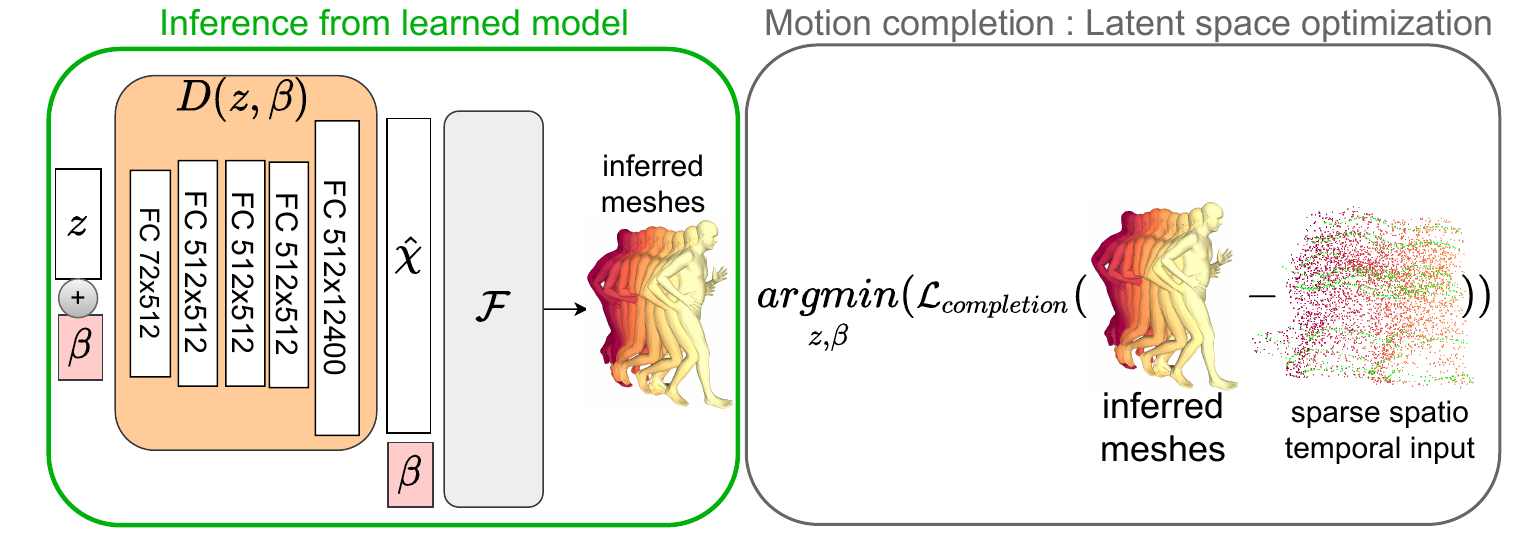}
 
    \caption{Motion completion. We minimize a loss \wrt~latent representation ($\latentVec, \morphology)$. Left: inference pipeline. Right: we optimize $\loss_{completion}$ between a sparse 4D point cloud and inferred meshes.}
    
    \label{fig:comp_method}
\end{figure}

We consider as input partial motion sequences of unordered dense 3D scans with possibly additional synchronized MoCap for $k$ landmarks and associated time stamps. Let $S=[s_1,\ldots,s_n]$ denote a sequence of $n$ anchor scans uniformly sampled in time, $L = [l_1,\ldots,l_n]$ the corresponding synchronized sequence of landmarks, and $\timeSeq = [\timestamp_{1},\ldots,\timestamp_{n}]$ the corresponding time stamps. Some anchor frames are empty, and our input consists of a set $I$ of frame indices $i$ for which $s_i$ or $l_i$ and $\timestamp_i$ are given.

To compute a sequence of anchor meshes $\hat \meshSeq$ with associated time stamps $\hat \timeSeq$ that approximate the input, we decode a full sequence of anchor frames $[\hat \meshSeq, \hat \timeSeq]$ using $\smseq\left( D(\latentVec, \morphology), \morphology \right)$ and optimize for latent vectors  $\latentVec^*,\morphology^*$ as
{\footnotesize
\begin{equation}
\latentVec^*,\morphology^{*} = \underset{\latentVec,\morphology}{argmin}(\loss_{completion}(\hat{\meshSeq}(\latentVec,\morphology),\hat{\timeSeq}(\latentVec,\morphology),S,L,\timeSeq)),
\end{equation}}
where 
{\footnotesize
\begin{eqnarray}
    \loss_{completion} &=& \omega_{dense}\sum_{i\in I} \mbox{Chamfer}(\hat\mesh_i(\latentVec, \morphology), s_i) \nonumber \\
    &+& \omega_{mocap}\sum_{i\in I} \mbox{Landmark}(\hat\mesh_i(\latentVec, \morphology), l_i)\nonumber \\
    &+& \omega_{time}\sum_{i\in I}(\hat{\tau_i}(\latentVec, \morphology)-\tau_{i})^2.
\end{eqnarray}}
%
The weights $\omega_{dense}$, $\omega_{mocap}$ and $\omega_{time}$ are adaptive~\cite{pmlr-v80-chen18a}. When $s_i=\emptyset, \omega_{dense}=0$ and when $l_i=\emptyset, \omega_{mocap}=0$. Varying $\omega_{mocap}$ allows to evaluate the benefit of having tracked input markers. $\mbox{Chamfer}$ is the Chamfer distance between two point clouds and $\mbox{Landmark}$ is the squared Euclidean distance between $k$ vertices of the \sm~template, selected once for all experiments, and the $k$ given landmarks. This optimization is visualized in Fig.~\ref{fig:comp_method}.

\subsection{Completion dataset} 
\label{sec:completion_dataset}
We introduce a new dataset of cyclic human motion (CHUM), which was captured using a 4D modeling platform with 68 RGB cameras and a Qualisys MoCap system. Data consists of dense scans of approximately 10000 points acquired at 50fps with synchronised MoCap for 16 markers. We recorded 4 actors with different morphologies (2 males and 2 females) performing various cyclic motions like walking, running, side-stepping, skipping, boxing and kicking.
For our experiment, we segmented 4 gait cycles manually for each original sequence and found an initial 3D transformation (rotation + translation) to align each segment at $t=0$. We do not fit \sm~to the dense scans because $\loss_{completion}$ does not require correspondence information.

\subsection{Results} 

We compare our results to three state of the art approaches. The first performs static 3D completion per frame~\cite{zhou2020reconstructing}.  Due to its high computational complexity, we apply the static method to a subset of CHUM while other methods are applied to the full dataset. This method is only applicable for spatial completion where observations are available at every frame.
The second and third are motion spaces for sequences of fixed duration that can serve as prior~\cite{xu2021exploring,li2021taskgeneric}. Given a partial motion as input, we optimize a latent motion vector $z$, a morphology $\morphology$ and a set of per-frame translation parameters for \cite{xu2021exploring}, as global translation is not encoded in this motion space. In case of temporally sparse input, translation parameters are only optimized for frames in $I$ and the remaining are found using linear interpolation between the closest observed frames. For \cite{xu2021exploring} and \cite{li2021taskgeneric}, we optimize for $\loss_{completion}$ with $\omega_{time}=0$, as these motion spaces are designed for sequences of fixed duration and cannot benefit from time stamp information. These methods are applicable for both spatial and temporal completion.
\cite{xu2021exploring} uses a latent space of 256 dimensions while  \cite{li2021taskgeneric} uses a total of 36. For fair comparison to the more precise method, we re-train our model with $dim(z)=256$.

\begin{table}
\caption{Comparative evaluation of motion completion. Mean and standard deviation of Chamfer distance in $mm$, computed between completions and ground truth anchor scans from CHUM. N.A. means not applicable.
}

\label{tab:completion-spatial}
    \scalebox{0.6}{\begin{subtable}{\textwidth}
        \small{\begin{tabular}{|c||c|c|c|c|c||c|c|c|}
        \hline
        &\multicolumn{5}{c||}{Points per scan $p$}&\multicolumn{3}{c|}{Frames ($f$)}\\
        \hline
          & 0 & 50 & 100 & 1000 & 10000 & 5 & 20 & 100\\
        \hline
        Ours (dim(z)=256)  & \textbf{42$\pm$48} & \textbf{23$\pm$7} & \textbf{21 $\pm$9} & \textbf{20$\pm$10} & 20$\pm$10 &  30$\pm$14 & \textbf{20$\pm$10} & \textbf{20$\pm$10}\\
        \hline
        \cite{zhou2020reconstructing}  & N.A. & 58$\pm$0.99 & 47 $\pm$0.6 & 21$\pm$0.3 & \textbf{10$\pm$0.46} & N.A. & N.A. & N.A.\\
        \hline
        
        \cite{xu2021exploring}  & 46 $\pm$ 52 & 26$\pm$8 & 24 $\pm$9 & 22$\pm$10 & 22$\pm$10 & \textbf{22$\pm$10} & 22$\pm$11 & 22$\pm$10\\
        \hline
        \cite{li2021taskgeneric}  & 216 $\pm$ 41 & 88$\pm$10 & 69 $\pm$10 & 36$\pm$10 & 26$\pm$11 & 33$\pm$13 & 26$\pm$11 & 26$\pm$11\\
        \hline
        \end{tabular}}
    \end{subtable}}
    
\end{table}

\textbf{Spatial completion}
We first evaluate the quality of spatial completion by simulating different levels of spatial sparsity by varying the number of points $p$ per scan $s_i$. The sampled points are not in correspondence over time. 
Table~\ref{tab:completion-spatial} shows the evolution of the reconstruction error in $mm$ when varying $p$. Our method outperforms the static method~\cite{zhou2020reconstructing} for very sparse scans ($p<100$), the two methods are on-par for denser scans ($p=1000$), and the static method outperforms our method for dense scans ($p=10000$). This quality on sparse scans is achieved because our model optimizes for all frames simultaneously, so few points per scan suffice to find a plausible solution. The static method deforms a template, and can capture higher levels of geometric detail for dense scans. 
Our method further outperforms state of the art motion spaces~\cite{xu2021exploring,li2021taskgeneric}, in spite of being trained on significantly less motion data (4.5h for ours vs. 34h for~\cite{xu2021exploring,li2021taskgeneric}). Qualitative results are shown in supplementary material.

\textbf{Temporal completion}
Second, we evaluate the quality of temporal completion by varying the number of observed frames. 
To vary this number for each test sequence, we reduce $I$ to simulate lower frame rates. Table~\ref{tab:completion-spatial} shows the evolution of the reconstruction error. The $f=100$ frame completion task includes all frames and is given as reference. The model extrapolates with almost no loss of precision with $I_{20}=[5,10,\ldots,95,100]$ (20 frames) and the error is still low with $I_5=[20,40,60,80,100]$ (5 frames). While the motion space for sequences of fixed duration~\cite{xu2021exploring} is better for sparsely sampled temporal data, we outperform both \cite{xu2021exploring} and \cite{li2021taskgeneric} for temporally denser data, in spite of using significantly less training data. 

\section{Conclusions and future work}

This work presents a latent space that allows to represent and generate multi-frame sequences of human motion in 4D. This latent space contains information on global motion, body pose, temporal evolution of the motion, and morphology. We demonstrated that similar motions tend to form clusters in this latent space and that linear interpolations between pairs of sequences in latent space are plausible. Furthermore, our model to generate 4D motion sequences captures the interaction between morphology and motion. We applied this model to spatio-temporal motion completion, demonstrating state of the art performance.
For future work, it would be interesting to explore how to synthesize longer term and more general motion. 

\section{Acknowledgements}

We thank Jinlong Yang and Jiabin Chen for the Kinovis test set, Joao Regateiro, Anne-H\'{e}l\`{e}ne Olivier and Edmond Boyer for helpful discussions, and Laurence Boissieux, Julien Pansiot, and our volunteer subjects for help with 4D data acquisition. This work was supported by the National French Research Agency under grants 19-CE23-0013-01 (3DMOVE) and ANR-21-ESRE-0030 (CONTINUUM). 

{\small
\bibliographystyle{ieee_fullname}
\bibliography{refs}
}



\section*{Supplementary material}

# Supplementary Material for *A Structured Latent Space for Human Body Motion Generation*

Mathieu Marsot[1]  Stefanie Wuhrer[1]  Jean-Sébastien Franco[1]  Stephane Durocher[2]

[1] Univ. Grenoble Alpes, Inria, CNRS, Grenoble INP[*], LJK, 38000 Grenoble, France

[2] University of Manitoba 66 Chancellors Cir, Winnipeg MB R3T 2N2, Canada

firstname.lastname@inria.fr, stephane.durocher@umanitoba.ca

## 1. Data

We start by providing more details on the training and test data used from the AMASS [3] and Kinovis [6] datasets. AMASS regroups a large set of MoCap recordings and fits a parametric body model to all data. When splitting the AMASS dataset into training and test sets, we treat all sequences emanating from the same MoCap dataset as one entity. We leave the MoCap datasets "MPI_mosh", "SFU", and "TotalCapture" for testing, and call this dataset *AMASS test set*. The Kinovis dataset contains 4D motion sequences captured using a multi-view platform and allows to evaluate the generalization of the model to densely captured 4D data. We consider all walking and running sequences, pre-process the data by fitting SMPL before extracting cyclic hip motions, and call this dataset *Kinovis test set*.

**Training data** We automatically extract 12085 sequences of motion cycles with various duration and motion types from the AMASS training set which amounts to approximately 4.5 hours of motion data. To allow for efficient learning, the sequences are spatially aligned by zeroing the initial translation and we use the identity rotation as initial rotation of the root joint to be invariant in the ground plane.

**Test data** We consider two test datasets. The first one is called *AMASS test set* in the following and contains 1027 sequences extracted from the AMASS test set which amounts to approximately 20 minutes of motion data. The second one, called *Kinovis test set* in the following, contains 4D motion sequences captured using a multi-view platform. This dataset is an extension of [6] and allows to evaluate the generalization of the model to densely captured 4D data. We consider all walking and running sequences, and pre-process the data by fitting SMPL to the 4D sequences before extracting cyclic hip motions. This results in 37 test sequences, some of which contain less than 100 frames; we augment shorter sequences to 100 frames using linear interpolation between the 6D rotations.

## 2. Implementation Details

In our motion representation $\chi$, we do not consider the SMPL components related to hands or dynamic components available in AMASS. We further discard the two foot joints because they have constant rotation. This leaves a total of 20 joints. Our representation $\chi$ consists of 100 timestamped anchor meshes, each of which is represented by 124 parameters (120 for $\theta$, 3 for $\gamma$ and 1 for $\tau$). 100 anchor meshes are chosen as they provide a good trade-off between the error introduced by the sampling and the dimensionality of $\chi$. To normalize the data, we normalize the translation $\gamma$ in $[-1, 1]^3$, and the timestamps $\tau$ in $[0, 1]$ using minmax scaling over the training set. We remove the identity rotation $[1, 0, 0, 0, 1, 0]$ from the 6D representation, which leads to a significant gain in reconstruction accuracy compared to the classic scaling $\frac{\theta - \mu_\theta}{\sigma_\theta}$.

We train for 5000 epochs with $\mathcal{L}_{init}$, using a learning rate of $1e^{-3}$ and a batch size of 256. Each epoch takes 6 s for a total training time of 8 hours. We train with $\mathcal{L}$ for 200 epochs using a smaller batch size of 16 for memory reasons and a learning rate of $1e^{-4}$. Here epoch time is 8 min for a total training time of one day. The training is done on a NVIDIA Quadro RTX8000 with 48G of GPU RAM. We use $\omega_{kl} = 0.01$ for both steps and chose a latent dimension for the motion space $z$ of 64, and of 8 for $\beta$. Note that mesh vertex positions are in meters during training. We initialize all dynamic weights to 1.0 and GradNorm [1] updates the weights dynamically.

## 3. Influence of latent space dimension and regularisation

We now investigate the influence of the latent space dimension and regularization on the quality of the model. To evaluate the model's quality, we measure its reconstruction

---

[*] Institute of Engineering Univ. Grenoble Alpes

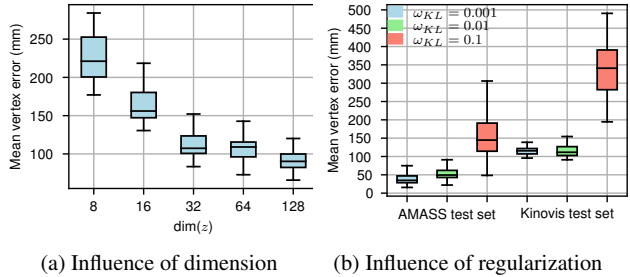

(a) Influence of dimension  (b) Influence of regularization

Figure 1: Influence of latent space dimension and regularisation on reconstruction error. Left: Increasing dimension of the latent space leads lower reconstruction errors on AMASS test set. Right: Smaller regularization $\omega_{KL}$ leads lower reconstruction errors on both test sets. Boxes follow [4].

error, which characterizes the model's ability to reconstruct examples unseen during training and is defined as

$$\frac{1}{nk}\left(M - \hat{M}\right)^2, \quad (1)$$

with $\left[\hat{M}, \hat{\mathcal{T}}\right] = \mathcal{F}(\hat{\chi}, \beta) = \mathcal{F}(D(E(\chi, \epsilon), \beta), \beta)$, where $n$ is the number of anchor frames and $k$ the number of vertices per frame. As second qualitative error measure, we consider the model's ability to allow for the generation of plausible new sequences by sampling in latent space. In practice, we consider samples that are linearly interpolated between sequences of the test set.

**Latent space dimension** We first study the influence of the dimensionality of the latent space on the model quality. Fig. 1a shows the impact of the dimension of $z$ on the reconstruction error on the AMASS test set. As expected, the bigger the latent space dimension, the smaller the error. However, for $dim(z) > 64$, the error starts to stagnate. Therefore we set the dimension of the latent space to $64$.

**Latent space regularisation** The regularisation of the latent space has a major impact on the model quality. It is controlled by coefficient $\omega_{KL}$, which weighs the influence of latent space regularization at the cost of reconstruction accuracy. Fig. 1b shows the reconstruction error on models trained with different values for $\omega_{KL}$. The smaller $\omega_{KL}$, the smaller the reconstruction error. However, with $\omega_{KL} = 0.001$, the model no longer allows generating plausible new sequences and a problematic interpolation is shown in supplementary material. Therefore, we set $\omega_{KL} = 0.01$ in the following.

## 4. Motion completion from spatially sparse input

A qualitative comparison for the completion task with $p = 100$ and available landmark data is shown in Fig. 2. Note that our method leads to more plausible wrist and hand motion than [5], better temporal coherence than [7], better leg motion than [7] and [2] and better global translation than [2].

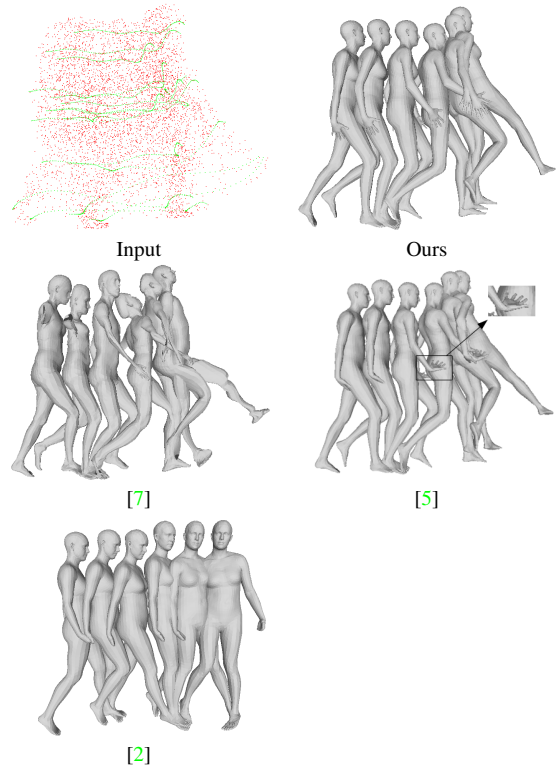

Figure 2: Qualitative comparison of spatial completion on kick sequence from CHUM with $p = 100$. Input scans shown in red, landmarks in green. Visualization shows 6 of 100 completed frames. Note that our motion completion is plausible and coherent with input.



\end{document}